\documentclass[letterpaper]{article} %
\usepackage{aaai23}  %
\usepackage{times}  %
\usepackage{helvet}  %
\usepackage{courier}  %
\usepackage[hyphens]{url}  %
\usepackage{graphicx} %
\usepackage{natbib}  %
\usepackage{caption} %
\usepackage{algorithm}
\usepackage{algorithmic}

\usepackage{newfloat}
\usepackage{listings}
\DeclareCaptionStyle{ruled}{labelfont=normalfont,labelsep=colon,strut=off} %
\floatstyle{ruled}
\newfloat{listing}{tb}{lst}{}
\floatname{listing}{Listing}
\usepackage{amsmath,amsfonts,amssymb}
\usepackage{multirow}
\usepackage{booktabs}
\usepackage{paralist}
\usepackage{enumitem}
\usepackage{cleveref}
\usepackage{caption}
\usepackage{subcaption}
\usepackage{bm}
\usepackage{rotating}
\usepackage[mathscr]{euscript}
\usepackage{makecell}
\usepackage{colortbl}
\usepackage{adjustbox}

\usepackage{CJKutf8}

\newcommand{\abr}[1]{\textsc{#1}}
\newcommand{\mbf}[1]{\bm{\mathbf{#1}}}
\newcommand{\mscript}[1]{\text{\scriptsize{#1}}}
\crefname{equation}{Eq.}{Eq.}
\crefname{section}{Sec.}{Sec.}

\title{
    InfoCTM: A Mutual Information Maximization Perspective of \\ Cross-Lingual Topic Modeling
}

\author{
    Xiaobao Wu\textsuperscript{\rm 1}, \;
    Xinshuai Dong\textsuperscript{\rm 2}, \;
    Thong Nguyen\textsuperscript{\rm 3}, \;
    \\
    Chaoqun Liu\textsuperscript{\rm 1,4}, \;
    Liang-Ming Pan\textsuperscript{\rm 3}, \;
    Anh Tuan Luu\textsuperscript{\rm 1}
}
\affiliations{
    \textsuperscript{\rm 1}Nanyang Technological University, Singapore \\
    \textsuperscript{\rm 2}Carnegie Mellon University, USA \\
    \textsuperscript{\rm 3}National University of Singapore, Singapore \\
    \textsuperscript{\rm 4}DAMO Academy, Alibaba Group, Singapore \\
    \{xiaobao002,chaoqun001,anhtuan.luu\}@ntu.edu.sg\;
    xinshuad@andrew.cmu.edu\;
    \{e0998147,liangmingpan\}@u.nus.edu

}

\begin{document}

\maketitle

\begin{abstract}
    Cross-lingual topic models have been prevalent for cross-lingual text analysis by revealing aligned latent topics.
    However, most existing methods suffer from producing repetitive topics that hinder further analysis and performance decline caused by low-coverage dictionaries.
    In this paper, we propose the Cross-lingual Topic Modeling with Mutual Information (InfoCTM).
    Instead of the direct alignment in previous work, we propose a topic alignment with mutual information method.
    This works as a regularization to properly align topics and prevent degenerate topic representations of words, which mitigates the repetitive topic issue.
    To address the low-coverage dictionary issue, we further propose a cross-lingual vocabulary linking method
    that finds more linked cross-lingual words for topic alignment beyond the translations of a given dictionary.
    Extensive experiments on English, Chinese, and Japanese datasets demonstrate that
    our method outperforms state-of-the-art baselines, producing more coherent, diverse, and well-aligned topics and showing better transferability for cross-lingual classification tasks.
\end{abstract}

\section{Introduction}

\begin{figure}[!t]
    \centering
    \includegraphics[width=\linewidth]{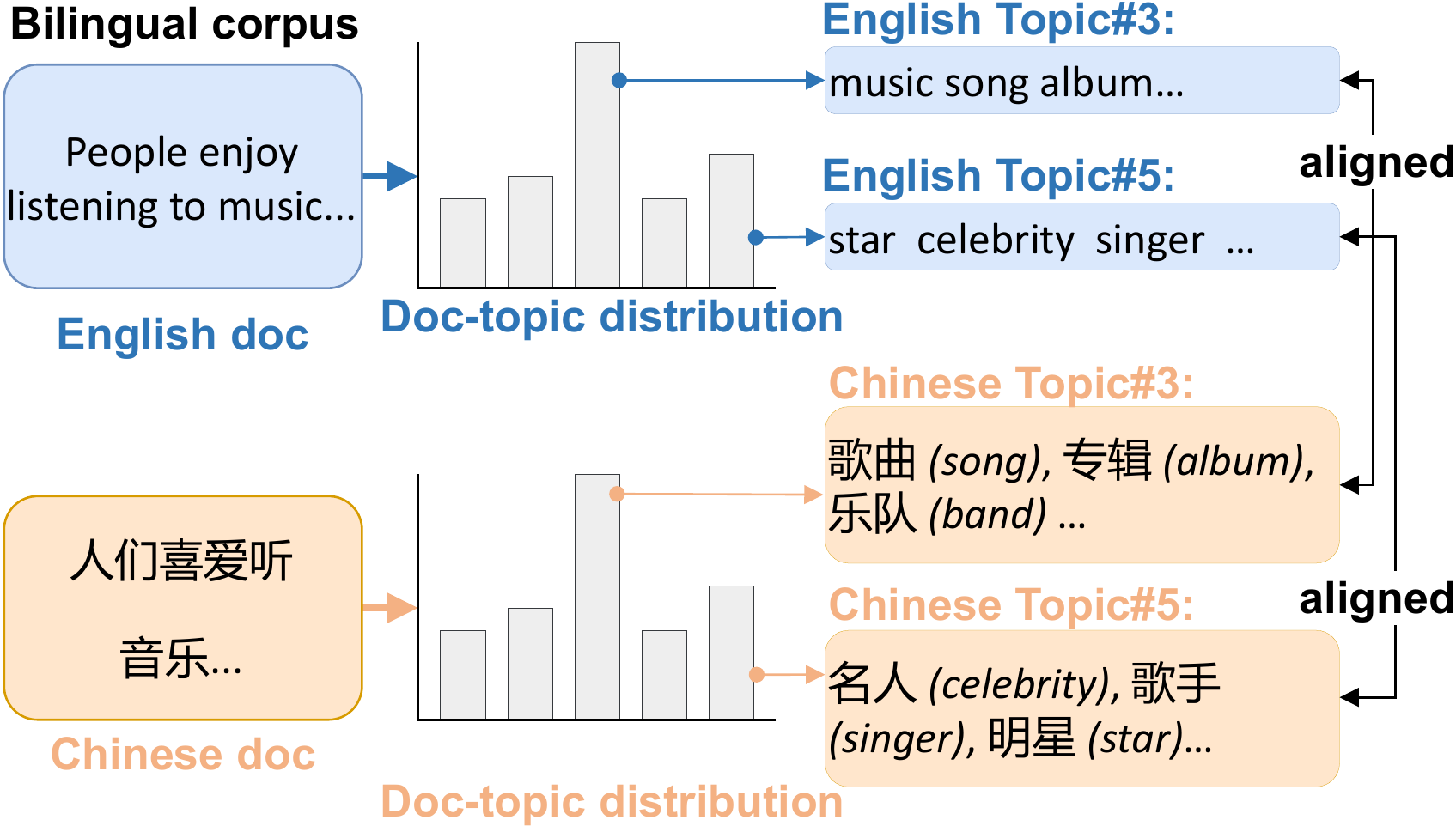}
    \caption{
        Illustration of cross-lingual topic models, producing aligned topics of different languages.
        \emph{Words} in the brackets are the corresponding English translations.
    }
    \label{fig_crosslingual_example}
\end{figure}

    Cross-lingual topic models have been popular for cross-lingual text analysis and applications \cite{vulic2013cross}.
    As shown in \Cref{fig_crosslingual_example},
    they aim to discover cross-lingual topics from bilingual corpora.
    Each topic is interpreted as the relevant words in the corresponding language.
    The same cross-lingual topics are required to be aligned (semantically consistent across languages).
    For example,
    the English Topic\#3 and Chinese Topic\#3 are aligned as they are both about music,
    and English Topic\#5 and Chinese Topic\#5 are aligned and both about the celebrity.
    These aligned topics can reveal commonalities and differences across languages and cultures,
    which enables cross-lingual analysis without supervision
    \cite{ni2009mining,shi2016detecting,gutierrez2016detecting,lind2019bridge}.

    Since parallel corpora are often difficult to access,
    recent cross-lingual topic models tend to rely on vocabulary linking information from bilingual dictionaries \cite{shi2016detecting,yuan2018multilingual,yang2019multilingual,Wu2020}.
    They commonly use translations of a dictionary as linked cross-lingual words
    and make these words belong to the same cross-lingual topics, \textit{i.e.}, align their topic representations (what topics a word belongs to).
    For instance in \Cref{fig_crosslingual_example}, the word ``song'' and its Chinese translation both belong to Topic\#3 of English and Chinese.
    These methods are more practical because dictionaries are widely accessible.
    Recent studies \cite{bianchi2020cross,mueller2021fine} employ multilingual BERT \cite{devlin2018bert} for multilingual corpora,
    but they are not traditional cross-lingual topic models since they do not discover aligned cross-lingual topics.

\begin{CJK*}{UTF8}{gbsn}
    \begin{table}[!t]
        \centering
        \small
        \setlength{\tabcolsep}{1.5pt}
            \begin{tabular}{rcccccccc}
                \toprule
    
                \textbf{English Topic\#1:} & \underline{photos} & \underline{style} & finds & ebay & \underline{week} & vintage \\
                \textbf{Chinese Topic\#1:} & 风格 & 文字 & 西装 & \underline{模特} & 西服 & \underline{每周} \\
    
                \midrule
    
                \textbf{English Topic\#2:} & \underline{fashion} & \underline{week} & \underline{photos} & new & \underline{style} & \underline{beauty} \\
                \textbf{Chinese Topic\#2:} & \underline{时尚} & \underline{时髦} & \underline{流行} & \underline{时装} & \underline{模特} & \underline{每周}\\
    
                \midrule
    
                \textbf{English Topic\#3:} & \underline{photos} & \underline{fashion} & new & \underline{beauty} & line & \underline{week} \\
                \textbf{Chinese Topic\#3:} & \underline{时尚} & \underline{时髦} & 全新 & 金山 & \underline{流行} & \underline{模特} \\
    
                \bottomrule
            \end{tabular}%
        \caption{
            Top related words of of repetitive cross-lingual topics produced by MCTA~\cite{shi2016detecting}.
            Repetitive words are underlined.
        }
        \label{tab_motivation}
    \end{table}
    \end{CJK*}

    However,
    despite the practicality,
    these methods, \textit{e.g.}, MCTA \cite{shi2016detecting} and NMTM \cite{Wu2020}, suffer from two issues:
    \begin{inparaenum}[(\bgroup\bfseries 1\egroup)]
        \item
            They tend to generate low-quality \emph{repetitive cross-lingual topics}, as exemplified in Table~\ref{tab_motivation}. 
            We see they all refer to similar semantics with many repetitive words like ``fashion'' and ``photo''.
            Consequently, this makes the discovered topics less useful for further text analysis and also hampers the performance of downstream applications.
        \item
            These methods mostly suffer from performance decline caused by \emph{low-coverage dictionaries}.
            Due to cultural differences,
            available bilingual dictionaries can only cover a small part of the involved vocabulary,
            especially for low-resource languages \cite{chang2021word}.
            Low-coverage dictionaries have been shown to hinder the topic alignment of cross-lingual topic models \cite{jagarlamudi2010extracting,hao2020empirical}.
            For example, it will be difficult to align English Topic\#3 and Chinese Topic\#3 in \Cref{fig_crosslingual_example} if we are unaware of the Chinese translations of English words like ``song'' or ``album''.
    \end{inparaenum}

    To address the above problems,
    we in this paper propose a novel neural cross-lingual topic model, named
    \textbf{C}ross-lingual \textbf{T}opic \textbf{M}odeling with Mutual \textbf{Info}rmation (InfoCTM).
    First,
        to address the repetitive topic issue,
        we propose a Topic Alignment with Mutual Information (TAMI) method.
        Instead of the direct alignment in previous work \cite{shi2016detecting,yang2019multilingual,Wu2020},
        TAMI maximizes the mutual information between the topic representations (what topics a word belongs to) of linked cross-lingual words.
        This not only aligns the topic representations of linked words
        but also prevents them from degenerating into similar values,
        which encourages words to belong to different topics.
        As a result,
        the discovered topics are distinct from each other, which alleviates the repetitive topic issue and enhances topic coherence and alignment.

    Second,
        to find linked words for TAMI and to overcome the low-coverage dictionary issue,
        we propose a Cross-lingual Vocabulary Linking (CVL) method.
        Instead of only using the translations in a dictionary as linked words,
        CVL additionally links a word to the translations of its nearest neighbors in the word embedding space.
        This is motivated by the fact that topic models focus on what topics a word belongs to rather than accurate translations. 
        For instance in \Cref{fig_crosslingual_example}, the English word ``album'' and the Chinese translation of ``song'' both belong to Topic\#3 of English and Chinese
        although they are not translations of each other.
        With CVL, we can obtain more linked cross-lingual words for our TAMI beyond the given dictionary,
        which mitigates the low-coverage dictionary issue.

    The contributions of this paper can be summarized as \footnote{Our code is available at \url{https://github.com/bobxwu/InfoCTM}.}:
    \begin{itemize}
        \item
            We propose a novel neural cross-lingual topic model with a new topic alignment with mutual information method that can prevent degenerate topic representations and avoid generating repetitive topics.
        \item
            We further propose a novel cross-lingual vocabulary linking method,
            which finds more linked cross-lingual words beyond the translations and effectively alleviates the low-coverage dictionary issue.
        \item
            We conduct extensive experiments on datasets of different languages
            and show that our model consistently outperforms baselines, producing higher-quality topics and showing better cross-lingual transferability for downstream tasks.
    \end{itemize}

\section{Related Work}
    \paragraph{Cross-lingual Topic Models}
        Cross-lingual topic modeling is proposed as an extension of monolingual topic modeling \cite{blei2003latent,blei2006dynamic,Wu2019}.
        The earliest polylingual topic model \cite{mimno2009polylingual} uses one topic distribution to generate a tuple of comparable documents in different languages, \emph{e.g.}, EuroParl \cite{koehn2005europarl}.
        As it is limited by the requirement of parallel/comparable corpus to link documents,
        another line of work uses vocabulary linking from bilingual dictionaries \cite{jagarlamudi2010extracting,boyd2012multilingual}.
        Recent studies of this line \cite{shi2016detecting,yang2019multilingual,Wu2020}
        commonly use translations in a dictionary as linked words
        and directly align topics by making these words belong to the same topics.
        \citet{chang2021word} induce translations by transforming cross-lingual word embeddings into the same space;
        however,
        they heavily rely on the isomorphism assumption \cite{conneau2017word} that cannot always hold as they find.
        Recently,
        \citet{bianchi2020cross,mueller2021fine} employ multilingual BERT \cite{devlin2018bert} to infer cross-lingual topic distributions for zero-shot learning,
        but they cannot discover aligned cross-lingual topics as required.
        Different from these work,
        we focus on two crucial issues of cross-lingual topic modeling: the repetitive topic issue and the low-coverage dictionary issue.
        To address these issues,
        we propose the topic alignment with mutual information instead of direct alignment
        and propose the cross-lingual vocabulary linking method instead of only using translations of a dictionary.
    
    \paragraph{Mutual Information Maximization}
        Mutual information maximization has been prevalent to learn visual and language representations \cite{bachman2019learning,kong2020,chi2020infoxlm,dong2021should}.
        In practice, mutual information maximization is approximated with a tractable lower bound, such as InfoNCE \cite{van2018representation} and InfoMax \cite{hjelm2018learning}.
        These are also known as contrastive learning \cite{arora2019theoretical,wang2020understanding,nguyen2022adaptive,Wu2022mitigating} that learns the representation similarity of positive and negative samples.
        Some recent studies \cite{xu2022neural} apply mutual information for monolingual topic modeling and focus on the representations of documents.
        We share the same perspective of information theory but look into a different problem, cross-lingual topic modeling.
        More importantly,
        instead of learning the representations of documents,
        we focus on the topic representations of words,
        which motivates our topic alignment with mutual information.
        This is also different from precedent work.

\section{Methodology}

    We first introduce the problem setting of cross-lingual topic modeling.
    Then, we present
    our new method Cross-lingual Topic Modeling with Mutual Information (InfoCTM), comprised of Topic Alignment with Mutual Information (TAMI) and Cross-lingual Vocabulary Linking (CVL).

    \subsection{Problem Setting and Notations}
        Consider a bilingual corpus of language $\ell_1$ and $\ell_2$ (\emph{e.g.}, English and Chinese).
        The vocabulary sets of each language are $\mathcal{V}^{(\ell_1)}$ and $\mathcal{V}^{(\ell_2)}$ with sizes as $V_1$ and $V_2$.
        Letting $w_i$ denote the $i$-th word type in the bilingual corpus,
        we assume the previous $V_1$ words are in language $\ell_1$ and the last $V_2$ words are in language $\ell_2$: $\mathcal{V}^{(\ell_1)} = \{w_i|i\!\!=\!\!1, \dots, V_1\}$ and $\mathcal{V}^{(\ell_2)} \! = \! \{w_i | i\!=\!V_1\!+\!1, \dots, V_1\!+\!V_2\}$.
        As illustrated in \Cref{fig_crosslingual_example},
        cross-lingual topic models aim to discover $K$ topics for each language from the bilingual corpus.
        Each topic of a language is defined as a distribution over words in the vocabulary (topic-word distribution).
        Namely, the Topic\#$k$ of language $\ell_1$ and $\ell_2$ are defined as
        $\mbf{\beta}_{k}^{(\ell_1)} \!\! \in \!\! \mathbb{R}^{V_1}$ and $\mbf{\beta}_{k}^{(\ell_2)} \!\! \in \!\! \mathbb{R}^{V_2}$ respectively.
        We require the Topic\#$k$ in language $\ell_1$ and the Topic\#$k$ in language $\ell_2$ to be aligned, \emph{i.e.}, semantically consistent across languages.
        For example, 
        the English Topic\#3 and Chinese Topic\#3 both focus on music in \Cref{fig_crosslingual_example}.
        Besides,
        cross-lingual topic models also infer what topics a document contains, \emph{i.e.}, the topic distributions of documents (doc-topic distributions), defined as $\mbf{\theta}^{(\ell_1)}, \mbf{\theta}^{(\ell_2)} \in \mathbb{R}^{K} $.
        We require the doc-topic distributions to be consistent across languages.
        If two documents in different languages contain similar topics,
        their inferred doc-topic distributions should also be similar.
        For instance, \Cref{fig_crosslingual_example} shows that the doc-topic distributions of the parallel English and Chinese document are similar.

\begin{figure}[!t]
    \centering
    \includegraphics[width=0.6\linewidth]{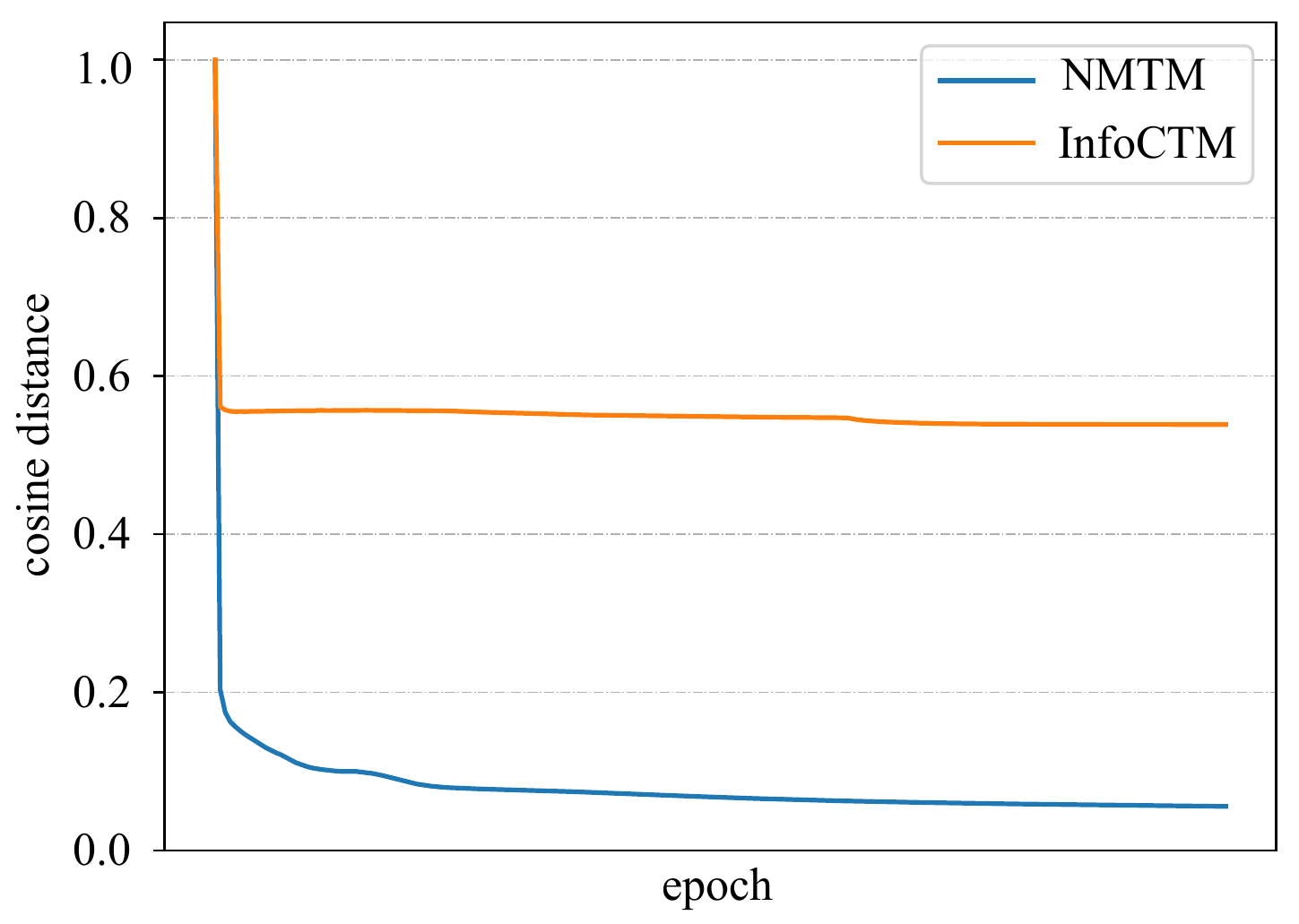}
    \caption{
        Cosine distance between the topic representations of words over the course of training.
        The results show that while the topic representations degenerate into similar values in NMTM \cite{Wu2020},
        our InfoCTM successfully avoids degenerate topic representations.
    }
    \label{fig_cosine_distance}
\end{figure}

    \subsection{Aligning Topics across Languages By Maximizing Mutual Information}
        We first analyze what causes repetitive topics with a state-of-the-art method, and then provide our solution called topic alignment with mutual information.

        \paragraph{What Causes Repetitive Topics?}
            In order to align topics,
            recent methods commonly use translations of a dictionary as linked cross-lingual words
            and directly align their topic representations.
            The topic representation of a word represents what topics this word belongs to.
            For example,
            \citet{yang2019multilingual} computes the topic distributions of words as topic representations and aligns them through inference,
            and \citet{shi2016detecting,Wu2020} transform topic representations of words to another vocabulary space and align them through generations.
            However, we find these methods using direct alignment have a severe issue:
            they easily produce repetitive topics (shown in \Cref{tab_motivation} and the experiment section).
            To investigate the behind reason,
            we compute the cosine distance between the learned topic representations in a state-of-the-art method, NMTM \cite{Wu2020}.
            \Cref{fig_cosine_distance} shows that
            the cosine distance is close to 0 after training in NMTM.
            This means NMTM ends with a trivial solution that all topic representations become similar.
            This is because the direct alignment of NMTM only encourages capturing the similarity between topic representations while ignoring the dissimilarity between them.
            As a result,
            all topic representations wrongly degenerate into similar values,
            and the discovered topics cover similar words, which leads to repetitive topics.

\begin{figure}[!t]
    \centering
    \includegraphics[width=\linewidth]{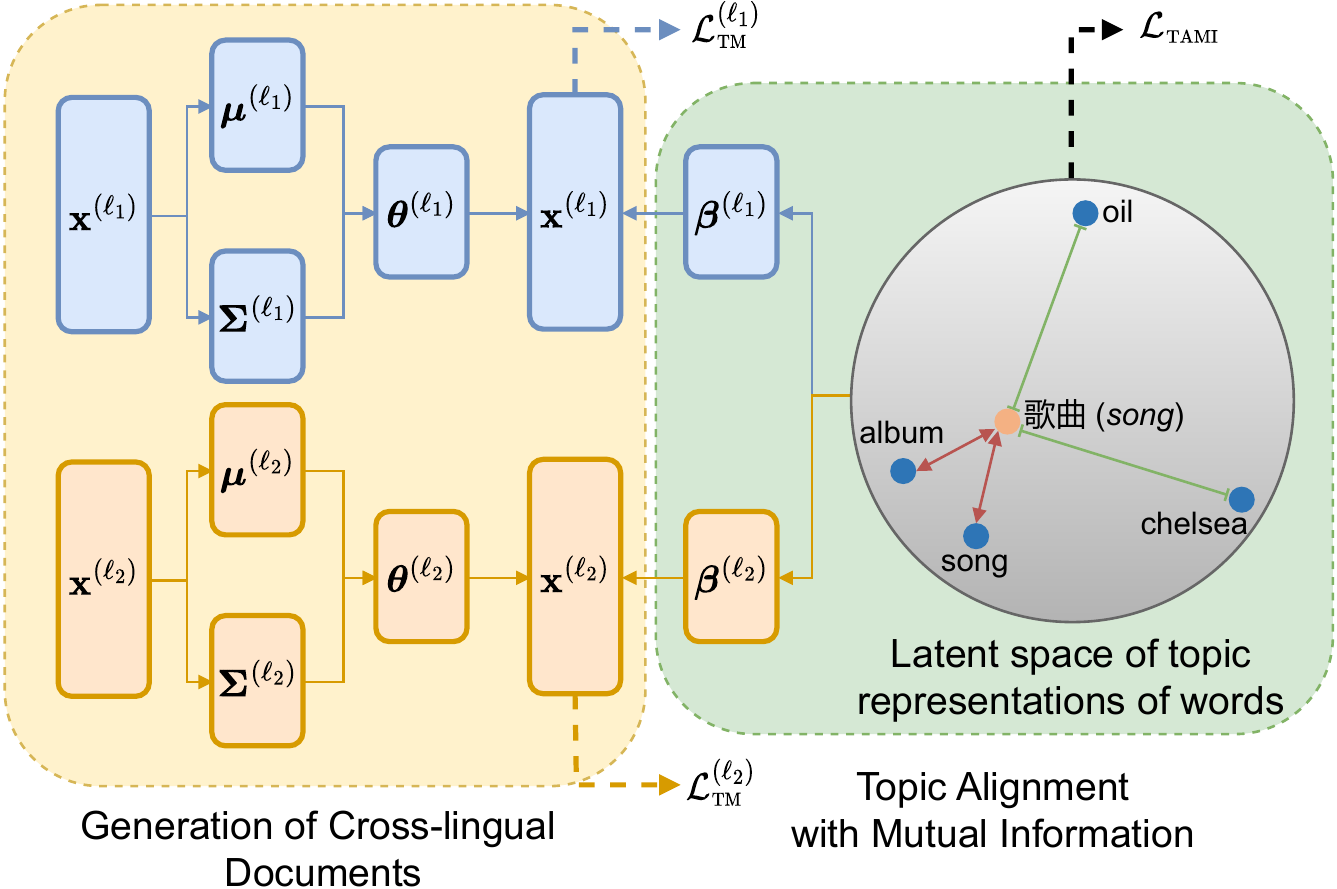}
    \caption{
        Illustration of InfoCTM.
        The generation of cross-lingual documents follows VAE.
        The proposed topic alignment with mutual information method aligns the topic representations of linked words
        (\begin{CJK*}{UTF8}{gbsn}``歌曲''\end{CJK*} (\emph{song}) and ``song'' or ``album'')
        and also
        keeps the distance between the topic representations of unlinked words
        (\begin{CJK*}{UTF8}{gbsn}``歌曲''\end{CJK*} (\emph{song}) and ``oil'' or ``chelsea'')
        to avoid degenerate topic representations.
    }
    \label{fig_model_illustration}
\end{figure}

        \paragraph{Topic Alignment with Mutual Information}
            Motivated by the above analyses,
            we aim to (i) capture the similarity between the topic representations of linked cross-lingual words
            and
            (ii) avoid degenerate topic representations.
            For these two purposes,
            we propose the topic alignment with mutual information (TAMI).
            \Cref{fig_model_illustration} illustrates the idea of our TAMI;
            \Cref{fig_cosine_distance} shows that our InfoCTM with TAMI can effectively avoid degenerate topic representations.
            Specifically,
            we define random variables $W$ and $W'$ as two linked cross-lingual words with related semantics, \emph{e.g.}, a translation pair.
            We can achieve the above two purposes by
            maximizing the mutual information between $W$ and $W'$ estimated by their topic representations:
            \begin{equation}
                \max \, I(W; W').
            \end{equation}
            Intuitively,
            this mutual information measures the dependency between $W$ and $W'$.
            Maximizing this dependency can make the topic representation of linked words similar.
            Meanwhile,
            this dependency is reduced if topic representations are all similar to each other since a word will be associated with every other word.
            Thus, maximizing this dependency can also keep the dissimilarity between the topic representations of unlinked words and thus avoid degenerate topic representations.

            Unfortunately, it is generally intractable to directly maximize mutual information when cooperating with neural networks,
            so we resort to a lower bound on it.
            One particular lower bound,
            InfoNCE \cite{logeswaran2018efficient,van2018representation} has been shown to work well in practice.
            Similarly,
            we relax the mutual information following InfoNCE as:
            \begin{align}
                &I(W; W') \geqslant \log |\mathscr{B}| + \notag \\
                & \mathbb{E}_{p(w_i, w_j)} \left[ \log \frac{\exp(g(f(w_i), f(w_j)) )}{\sum_{w_{j'} \in \mathscr{B}} \exp(g(f(w_i), f(w_{j'})) ) } \right]. \label{eq_bound}
            \end{align}
            Here $w_i$ and $w_j$ are specific values of $W$ and $W'$ respectively.
            $f: \mathcal{V}^{(\ell_1)} \cup \mathcal{V}^{(\ell_2)} \rightarrow \mathbb{R}^{K}$ denotes a lookup function that maps a word type $w_i$ into a vector $\mbf{\varphi}_i$ as its topic representation.
            So we have $g(f(w_i), f(w_j)) = g(\mbf{\varphi}_i, \mbf{\varphi}_j)$.
            Function $g(\cdot, \cdot)$ is a critic
            to
            characterize the similarity between $\mbf{\varphi}_i$ and $\mbf{\varphi}_j$.
            We implement $g$ as a scaled cosine function \cite{wu2018unsupervised}: $g(a, b) = \cos(a, b) / \tau$
            where $\tau$ is a temperature hyper-parameter.
            Set $\mathscr{B}$ includes positive sample $w_j$ and $(|\mathscr{B}| - 1)$ negative samples.
            This is also known as contrastive learning \cite{chen2020simple,tian2020contrastive}
            where we pull close the topic representations of a positive pair $(w_i, w_j)$ and push away the topic representations of negative pairs $(w_i, w_{j'})$~($j' \neq j$).
            From the perspective of contrastive learning, the maximization of mutual information can also be justified by the alignment and uniformity following \citet{wang2020understanding}.
            Maximizing the mutual information encourages the alignment and uniformity of topic representations of words in the latent space, and thus they are prevented to degenerate into close points.

    \paragraph{Cross-lingual Vocabulary Linking}
        Now we describe how to find linked cross-lingual word pair $(w_i, w_j)$ (a positive pair).
        As previous work \cite{shi2016detecting,yuan2018multilingual,Wu2020},
        $(w_i, w_j)$ can be a translation pair sampled from a bilingual dictionary.
        However,
        dictionaries could be low-coverage in real-world applications due to cultural differences, especially for low-resources languages.
        Low-coverage dictionaries provide insufficient translations and incur performance decline \cite{jagarlamudi2010extracting,hao2020empirical}.

\begin{figure}[!t]
    \centering
    \includegraphics[width=0.5\linewidth]{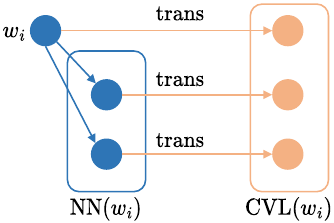}
    \caption{Illustration of Cross-lingual Vocabulary Linking.}
    \label{fig_vocab_linking}
\end{figure}

        To alleviate the low-coverage dictionary issue,
        we propose a cross-lingual vocabulary linking (CVL) method.
        CVL first prepares monolingual word embeddings via the commonly-used Word2Vec \cite{mikolov2013} for each language.
        As shown in \Cref{fig_vocab_linking},
        CVL then links word $w_i$ to the translations of its nearest neighbors in the embedding space besides its own translations.
        We denote $\mathrm{CVL}(w_i)$ as the linked word set of $w_i$, which is defined as
        \begin{equation}
           \mathrm{CVL}(w_i) = \bigcup_{w} \mathrm{trans}(w), \text{where} \; w \in \{w_i\} \cup \mathrm{NN}(w_i).
        \end{equation}
        Here,
        $\mathrm{NN}(w_i)$ denotes the set of nearest neighbors of word $w_i$.
        $\mathrm{trans}(w)$ denotes the translation set of word $w$ in a given dictionary.
        CVL views the cross-lingual words with related semantics as linked words instead of only translations.
        This is justified by the fact that topic modeling focuses on what topics a word belongs to rather than accurate translations.
        For example,
        English word ``album'' and the Chinese translation of ``song'' should belong to the same topic in English and Chinese although they are not translations of each other.
        Accordingly, our CVL method can easily infer more linked words beyond translations in a dictionary.

    \paragraph{Objective Function of Topic Alignment with Mutual Information}
        Let $N_{\mscript{CVL}}$ denote the number of all linked word pairs (positive pairs) found by the CVL method:
        $N_{\mscript{CVL}} = \sum_{i=1}^{V_1+V_2}|\mathrm{CVL}(w_i)|$.
        We then sample uniformly from all linked word pairs as
        $p(w_i, w_j) = \frac{1}{N_{\mscript{CVL}}}$ if $w_j \in \mathrm{CVL}(w_i)$ else $0$.
        Given a positive pair $(w_i, w_j)$,
        the negative samples  of $w_i$ in the set $\mathscr{B}$ are selected as the rest of words in the vocabulary set of $w_j$ except $\mathrm{CVL}(w_i)$:
        \begin{equation}
            \mathscr{B} = \{w_j\} \cup (\mathcal{V}^{(\ell)} \setminus \mathrm{CVL}(w_i))
        \end{equation}
        where $\ell$ refers to the language of $w_j$,
        and $\mathcal{V}^{(\ell)}$ is the vocabulary set of language $\ell$.
        Now we write the maximization of the lower bound (\Cref{eq_bound}) as minimizing $\mathcal{L}_{\mscript{TAMI}}$:
        \begin{equation}
            \! \mathcal{L}_{\mscript{TAMI}} \!\! = - \frac{1}{N_{\mscript{CVL}}} \!\!\! \sum_{i=1}^{V_1 + V_2} \!\!\!  \sum_{w_j \in \mathrm{CVL}(w_i)} \!\!\!\!\!\!\!\!\! \log \! \frac{\exp(g(\mbf{\varphi}_i, \mbf{\varphi}_j) )}{\sum_{w_{j'} \in \mathscr{B}} \exp(g(\mbf{\varphi}_i, \mbf{\varphi}_{j'}) ) }. \label{eq_MI}
        \end{equation}

\begin{table*}[!ht]
  \centering
  \small
  \setlength{\tabcolsep}{3mm}
  \renewcommand{\arraystretch}{1.05}
\begin{tabular}{lcccccccc}
\toprule
\multirow{2}[4]{*}{Model} & \multicolumn{2}{c}{EC News} &       & \multicolumn{2}{c}{Amazon Review} &       & \multicolumn{2}{c}{Rakuten Amazon} \\
\cmidrule{2-3}\cmidrule{5-6}\cmidrule{8-9}      & CNPMI & $TU$  &       & CNPMI & $TU$ &       & CNPMI & $TU$ \\
\midrule
MCTA  & 0.025$^\ddag$ & 0.489$^\ddag$ &       & 0.028$^\ddag$ & 0.319$^\ddag$ &       & 0.021$^\ddag$ & 0.272$^\ddag$ \\
MTAnchor & -0.013$^\ddag$ & 0.192$^\ddag$ &       & 0.028$^\ddag$ & 0.323$^\ddag$ &       & -0.001$^\ddag$ & 0.214$^\ddag$ \\
NMTM  & 0.031$^\ddag$ & 0.784$^\ddag$ &       & 0.042 & 0.732$^\ddag$ &       & 0.009$^\ddag$ & 0.679$^\ddag$ \\
\midrule
\textbf{InfoCTM} & \textbf{0.048} & \textbf{0.913} &       & \textbf{0.043} & \textbf{0.923} &       & \textbf{0.034} & \textbf{0.870} \\
\bottomrule
\end{tabular}%

    \caption{
        Topic quality results of topic coherence (CNPMI) and diversity ($TU$).
        The best are in bold.
        The superscript $\ddag$ means the improvements of InfoCTM is statistically significant at 0.05 level.
    }
    \label{tab_topic_quality}
\end{table*}%

    \subsection{Cross-lingual Topic Modeling with Mutual Information}
        In this section, we introduce InfoCTM by applying our proposed TAMI to the context of topic modeling through the generation of cross-lingual documents.
        \Cref{fig_model_illustration} illustrates the overall architecture of InfoCTM.

        \paragraph{Generation of Cross-lingual Documents}
            The generation process follows the framework of VAE \cite{Kingma2014a}
            as previous monolingual neural topic models \cite{Miao2016,Wu2020short,wu2021discovering,wu2022}.
            We use document $\mbf{x}^{(\ell_1)}$ in language $\ell_1$ to describe the generation process.
            First, we specify the prior and variational distribution.
            Following \citet{Srivastava2017},
            we use a latent variable $\mbf{r}^{(\ell_1)}$ with a logistic normal distribution as prior:
            $p(\mbf{r}^{(\ell_1)}) = \mathcal{LN}(\mbf{\mu}_{0}^{(\ell_1)}, \mbf{\Sigma}_{0}^{(\ell_1)}) $
            where $\mbf{\mu}_{0}^{(\ell_1)}$ and $\mbf{\Sigma}_{0}^{(\ell_1)}$ are the mean and the diagonal covariance matrix.
            The variational distribution is modeled as 
            $q_{\Theta_1}(\mbf{r}^{(\ell_1)} | \mbf{x}^{(\ell_1)}) = \mathcal{N}(\mbf{\mu}^{(\ell_1)}, \mbf{\Sigma}^{(\ell_1)})$ where $\mbf{\mu}^{(\ell_1)}$ and $\mbf{\Sigma}^{(\ell_1)}$ are the outputs of an encoder neural network with $\Theta_1$ as parameters.
            By applying the reparameterization trick \cite{Kingma2014a},
            we sample as $\mbf{r}^{(\ell_1)} = \mbf{\mu}^{(\ell_1)} + (\mbf{\Sigma}^{(\ell_1)})^{1/2} \mbf{\epsilon}$ where $\mbf{\epsilon} \sim \mathcal{N}(\mbf{0}, \mbf{I})$.
            The doc-topic distribution $\mbf{\theta}^{(\ell_1)}$ is modeled as 
            $\mbf{\theta}^{(\ell_1)} = \mathrm{softmax}(\mbf{r}^{(\ell_1)})$.

            To generate the document with $\mbf{\theta}^{(\ell_1)}$,
            we model the topic-word distribution matrices
            $\mbf{\beta}^{(\ell_1)} \!\! \in \!\! \mathbb{R}^{V_1 \times K}$ of language $\ell_1$ and
            $\mbf{\beta}^{(\ell_2)} \!\! \in \!\! \mathbb{R}^{V_2 \times K}$ of language $\ell_2$
            by the topic representations of words as:
            \begin{align}
                \mbf{\beta}^{(\ell_1)} &= (\mbf{\varphi}_1, \dots, \mbf{\varphi}_{V_1} )^{\top} \\
                \mbf{\beta}^{(\ell_2)} &= (\mbf{\varphi}_{V_1+1}, \dots, \mbf{\varphi}_{V_1+V_2} )^{\top}.
            \end{align}
            Then, we typically generate words in $\mbf{x}^{(\ell_1)}$ by sampling from a multinomial distribution:
            $
                x \!\! \sim \!\! \mathrm{Mult}(\mathrm{softmax}(\mbf{\beta}^{(\ell_1)} \mbf{\theta}^{(\ell_1)}))
            $
            \cite{Miao2016}.
            Similarly,
            the generation of document $\mbf{x}^{(\ell_2)}$ in language $\ell_2$ is formulated as $\mathrm{softmax}(\mbf{\beta}^{(\ell_2)} \mbf{\theta}^{(\ell_2)})$
            with parameter $\Theta_2$.

        \paragraph{Objective Function for Generation of Topic Modeling}
            Following the ELBO of VAE \cite{Kingma2014a},
            we formulate the generation objective of topic modeling as:
            \begin{align}
                \mathcal{L}_{\mscript{TM}}^{(\ell_1)} (\mbf{x}^{(\ell_1)}) = & - (\mbf{x}^{(\ell_1)})^{\top} \log ( \mathrm{softmax}(\mbf{\beta}^{(\ell_1)} \mbf{\theta}^{(\ell_1)})) \notag \\
                & + \mathrm{KL} \left[ q_{\Theta_1}(\mbf{r}^{(\ell_1)} | \mbf{x}^{(\ell_1)}) \| p(\mbf{r}^{(\ell_1)}) \right] . \label{eq_TM}
            \end{align}
            The first term measures the reconstruction error with $\mbf{x}^{(\ell_1)}$ in the form of Bag-of-Words as previous work \cite{Miao2016}.
            The second term is the KL divergence between the prior and variational distribution.
            Similar to \Cref{eq_TM}, we can easily write the objective function $\mathcal{L}_{\mscript{TM}}^{(\ell_2)}(\mbf{x}^{(\ell_2)})$ for document $\mbf{x}^{(\ell_2)}$.

        \paragraph{Overall Objective Function for InfoCTM}
            Letting $\mathcal{S}$ denote a set of cross-lingual documents,
            we write the overall objective function of InfoCTM with \Cref{eq_MI} and \Cref{eq_TM} as
            \begin{equation}
                \min_{\Theta_1, \Theta_2, \mbf{\varphi}} \!\!
                \lambda_{\mscript{TAMI}} \mathcal{L}_{\mscript{TAMI}} + \!\!\!\!\!\!\!\!\!\!
                \sum_{(\mbf{x}^{(\ell_1)}, \mbf{x}^{(\ell_2)}) \in \mathcal{S}} \!\!\!\!\!\!\!\!\!\!\!
                \frac{1}{|\mathcal{S}|} (\mathcal{L}_{\mscript{TM}}^{(\ell_1)} (\mbf{x}^{(\ell_1)}) + \mathcal{L}_{\mscript{TM}}^{(\ell_2)} (\mbf{x}^{(\ell_2)})) \notag
            \end{equation}
            where $\lambda_{\mscript{TAMI}}$ is a weight hyper-parameter.
            The $\mathcal{L}_{\mscript{TAMI}}$ objective works as a regularization of the generation objective of topic modeling,
            which aligns the topics across languages and meanwhile prevents degenerate topic representations.

\section{Experiment}
    In this section, we conduct extensive experiments to show the effectiveness of our method.
    \subsection{Experiment Setup}
        \paragraph{Datasets and Dictionaries}
            We use the following benchmark datasets in our experiments:
            \begin{itemize}[leftmargin=*]
                \item
                    \textbf{EC News}
                    is a collection of English and Chinese news \cite{Wu2020} with 6 categories: business, education, entertainment, sports, tech, and fashion.
                \item
                    \textbf{Amazon Review}
                    includes English and Chinese reviews from the Amazon website
                    where each review has a rating from one to five.
                    We simplify it as a binary classification task by labeling reviews with ratings of five as ``1''
                    and the rest as ``0'' following \citet{yuan2018multilingual}.
                \item
                    \textbf{Rakuten Amazon} contains Japanese reviews from Rakuten \cite[a Japanese online shopping website,][]{zhang2017encoding}, and English reviews from Amazon \cite{yuan2018multilingual}.
                    Similarly, it is also simplified as a binary classification task according to the rating.
            \end{itemize}
            We employ the entries from MDBG~\footnote{\url{https://www.mdbg.net/chinese/dictionary?page=cc-cedict}} as the Chinese-English dictionary for EC News and Amazon Review,
            and
            we use the Japanese-English dictionary from MUSE~\footnote{\url{https://github.com/facebookresearch/MUSE}} \cite{conneau2017word} for Rakuten Amazon.

        \paragraph{Baseline Models}
            We compare our method with the following state-of-the-art baseline models:
            \begin{inparaenum}[(i)]
                \item
                    \textbf{MCTA}~\cite{shi2016detecting},
                    a probabilistic cross-lingual topic model that detects cultural differences.
                \item
                    \textbf{MTAnchor}~\cite{yuan2018multilingual},
                    a multilingual topic model based on multilingual anchor words.
                \item
                    \textbf{NMTM}~\cite{Wu2020},
                    a neural multilingual topic model
                    which aligns topic representations by transforming them into the same vocabulary space.
            \end{inparaenum}
            We do not consider recent studies \cite{bianchi2020cross,mueller2021fine} because they do not discover aligned cross-lingual topics as required.

\begin{figure*}
    \centering
    \begin{subfigure}{0.333\linewidth}
        \centering
        \includegraphics[width=0.8\linewidth]{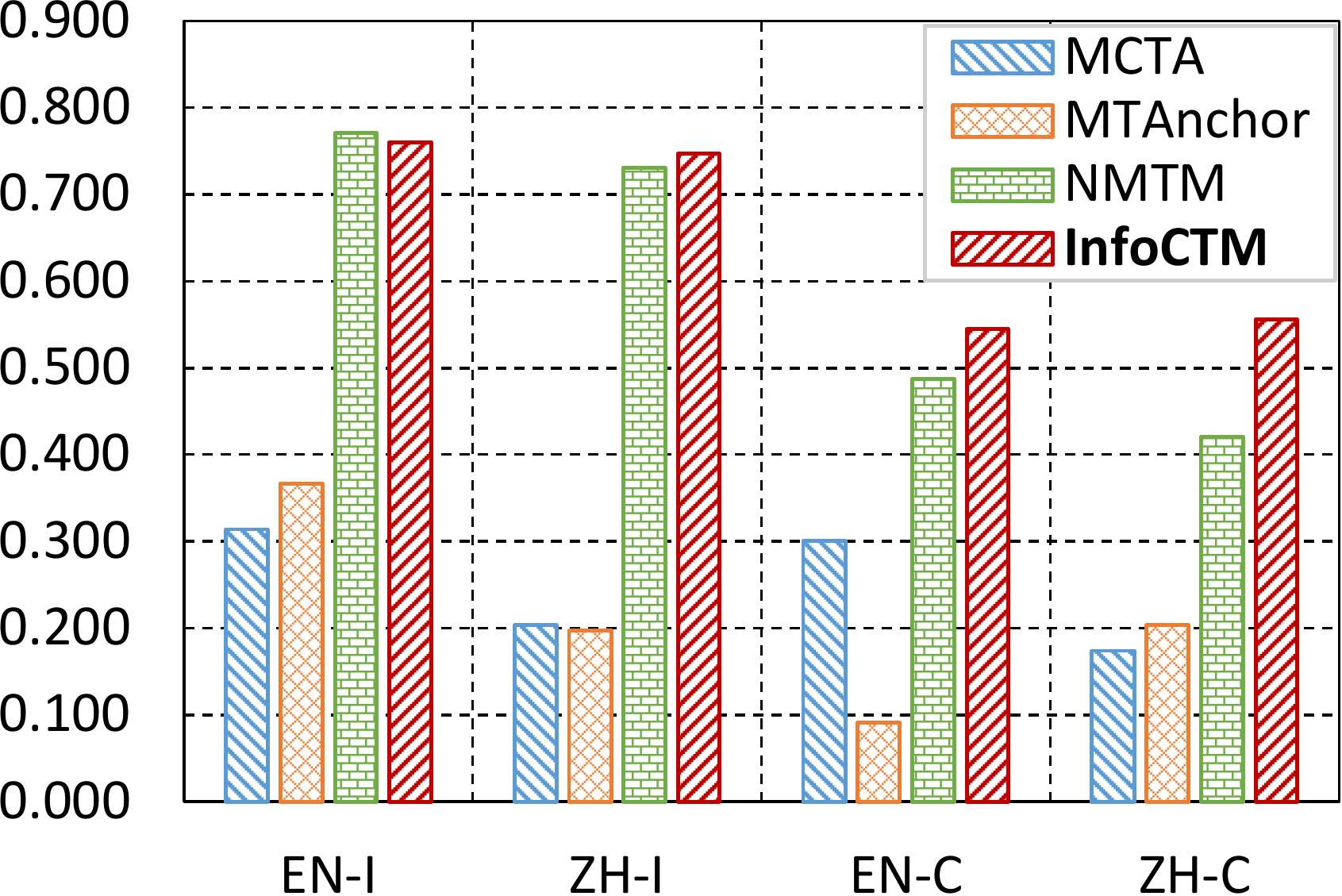}
        \caption{EC News}
        \label{fig_classification_ECNews}
    \end{subfigure}%
    \begin{subfigure}{0.333\linewidth}
        \centering
        \includegraphics[width=0.8\linewidth]{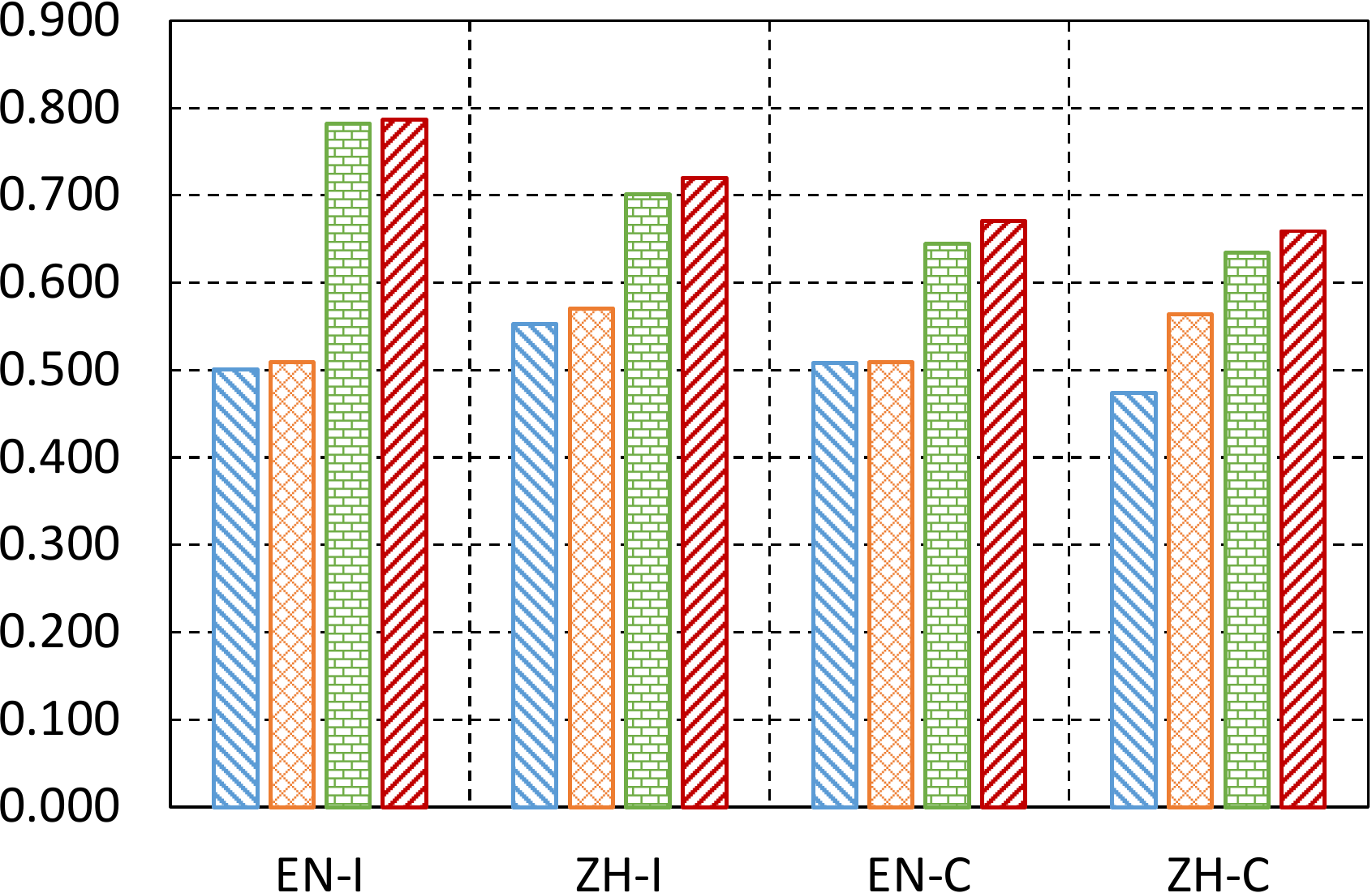}
        \caption{Amazon Review}
        \label{fig_classification_Amazon_Review}
    \end{subfigure}%
    \begin{subfigure}{0.333\linewidth}
        \centering
        \includegraphics[width=0.8\linewidth]{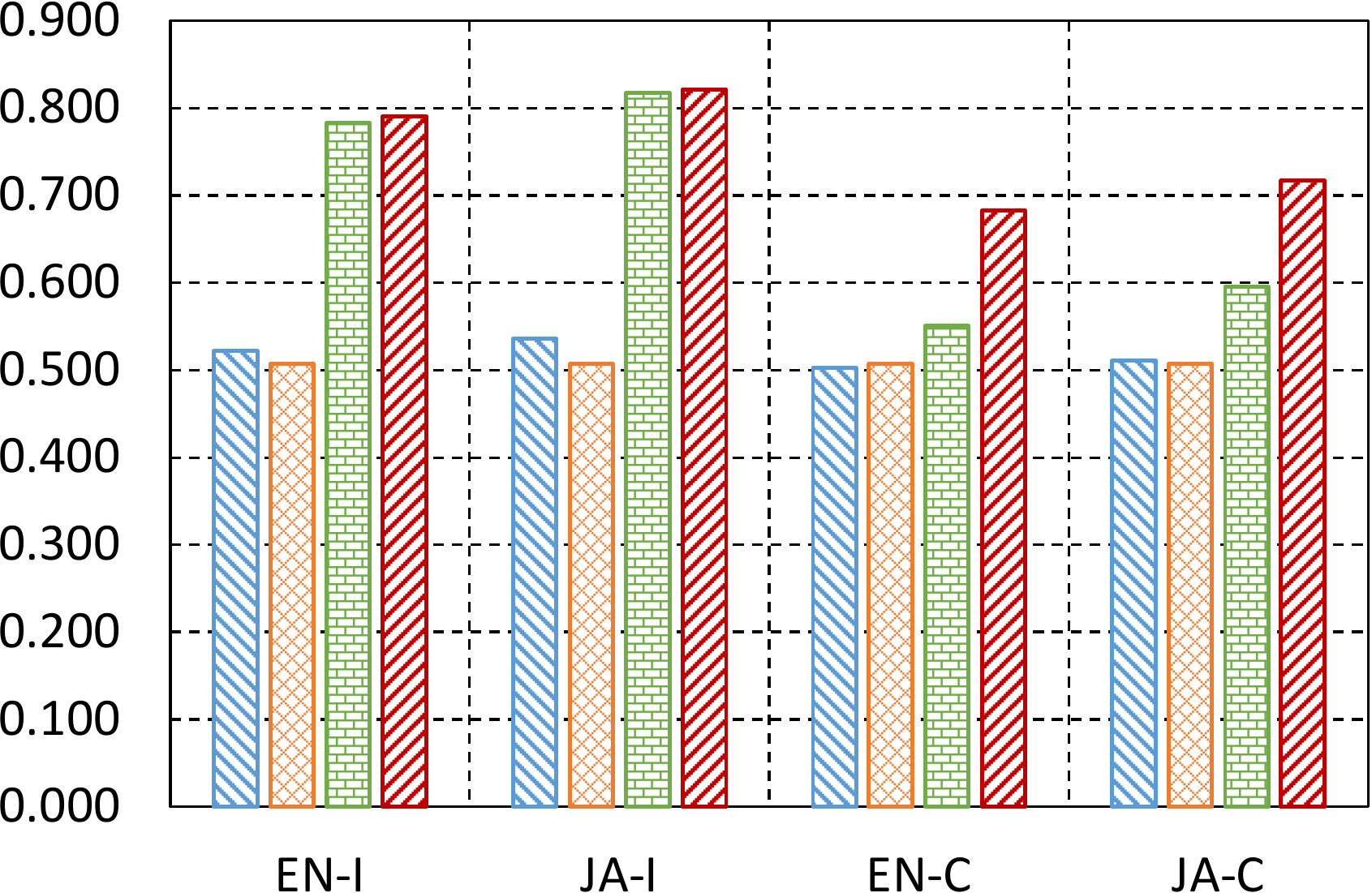}
        \caption{Rakuten Amazon}
        \label{fig_classification_Rakuten_Amazon}
    \end{subfigure}%

    \caption{
        Document classification accuracy
        where ``\abr{-i}'' means intra-lingual classification, and ``\abr{-c}'' is cross-lingual classification.
        Involved languages are English (\abr{en}), Chinese (\abr{zh}) and Japanese (\abr{ja}).
        The improvements of InfoCTM on cross-lingual classification (\abr{en-c,zh-c,ja-c}) are statistically significant at 0.05 level.
    }
    \label{fig_classification}
\end{figure*}

    \subsection{Cross-lingual Topic Quality} \label{sec_experiment_topic_quality}
        \paragraph{Evaluation Metrics}
            Following \citet{Wu2020,chang2021word},
            we evaluate topic quality from two perspectives:
            \begin{inparaenum}[(i)]
                \item 
                    \textbf{Topic coherence} evaluates the coherence and alignment of cross-lingual topics.
                    We use \textbf{CNPMI} \cite[Cross-lingual NPMI, ][]{hao2018lessons}, a popular metric for cross-lingual topics based on NPMI \cite{chang2009reading,Newman2010}.
                    CNPMI measures the coherence between the words in each topic of different languages,
                    \emph{e.g.}, between words in English Topic\#$k$ and words in Chinese Topic\#$k$.
                    Higher CNPMI indicates topics are more coherent and well-aligned across languages.
                \item
                    \textbf{Topic diversity} evaluates the difference between discovered topics to verify if they are repetitive.
                    We employ Topic Uniqueness ($TU$) \cite{Nan2019},
                    which calculates the proportion of different words in the discovered topics.
                    We report the average $TU$ of different languages for each dataset.
            \end{inparaenum}
            We select the top 15 related words of each topic for coherence and diversity evaluation.

        \paragraph{Result Analysis}
            \Cref{tab_topic_quality} summaries the topic coherence (CNPMI) and diversity ($TU$) results under 50 topics.
            We observe that baseline models generally suffer from repetitive topics: their $TU$ scores are quite low.
            As aforementioned, these repetitive topics are of low quality and can hinder further text analysis.
            In contrast, our InfoCTM consistently has much higher $TU$ under all the settings.
            \emph{E.g.}, InfoCTM achieves a $TU$ score of 0.913 on EC News while the runner-up is only 0.784.
            This is because InfoCTM adopts our topic alignment with mutual information,
            which prevents degenerate topic representations, alleviates repetitive topics, and thus improves topic diversity.
            In addition, InfoCTM achieves the best CNPMI scores on all datasets as in \Cref{tab_topic_quality}.
            For example, 
            InfoCTM has a CNPMI score of 0.034 on Rakuten Amazon, while the runner-up only has 0.021.
            Although on Amazon Review the CNPMI score of InfoCTM is only marginally larger than the runner-up, 
            we note the $TU$ of InfoCTM is much better (0.923 v.s. 0.732),
            and thus the overall topic quality of InfoCTM is higher.
            In summary, these results validate that InfoCTM can mitigate the repetitive topic issue and produce higher-quality cross-lingual topics than all the baselines.
            This advantage is crucial for further cross-lingual text analysis and applications.

\begin{table}[!t]
    \centering
    \small
    \setlength{\tabcolsep}{1.8pt}
    \renewcommand{\arraystretch}{1.2}
    \begin{tabular}{rlcccccc}
    \toprule
    \multicolumn{1}{c}{\multirow{2}[4]{*}{\makecell{Dict \\ Size}}} & \multicolumn{1}{c}{\multirow{2}[4]{*}{Model}} & \multicolumn{2}{c}{Topic Quality} & \multicolumn{4}{c}{Classification} \\
    \cmidrule{3-8}      &       & CNPMI & $TU$  & EN-I  & ZH-I  & EN-C  & ZH-C \\
    \midrule
    \multirow{3}[2]{*}{25\%} & NMTM  & 0.019$^{\ddag}$ & 0.763$^{\ddag}$ & 0.775 & 0.733 & 0.351$^{\ddag}$ & 0.348$^{\ddag}$ \\
          & w/o CVL & 0.035 & 0.795$^{\ddag}$ & \textbf{0.778} & \textbf{0.763} & 0.403$^{\ddag}$ & 0.356$^{\ddag}$ \\
          & \textbf{InfoCTM} & \textbf{0.036} & \textbf{0.895} & 0.769 & 0.755 & \textbf{0.472} & \textbf{0.448} \\
    \midrule
    \multirow{3}[2]{*}{50\%} & NMTM  & 0.025$^{\ddag}$ & 0.789$^{\ddag}$ & \textbf{0.775} & 0.730 & 0.403$^{\ddag}$ & 0.401$^{\ddag}$ \\
          & w/o CVL & \textbf{0.041} & 0.862$^{\ddag}$ & 0.772 & \textbf{0.753} & 0.433$^{\ddag}$ & 0.449$^{\ddag}$ \\
          & \textbf{InfoCTM} & 0.040 & \textbf{0.905} & 0.765 & 0.746 & \textbf{0.490} & \textbf{0.520} \\
    \midrule
    \multirow{3}[2]{*}{75\%} & NMTM  & 0.029$^{\ddag}$ & 0.803$^{\ddag}$ & \textbf{0.776} & 0.731 & 0.479$^{\ddag}$ & 0.441$^{\ddag}$ \\
          & w/o CVL & \textbf{0.045} & 0.884$^{\ddag}$ & 0.767 & 0.743 & 0.476$^{\ddag}$ & 0.462$^{\ddag}$ \\
          & \textbf{InfoCTM} & \textbf{0.045} & \textbf{0.909} & 0.761 & \textbf{0.748} & \textbf{0.519} & \textbf{0.537} \\
    \midrule
    \multirow{3}[2]{*}{100\%} & NMTM  & 0.031$^{\ddag}$ & 0.784$^{\ddag}$ & \textbf{0.771} & 0.731 & 0.487$^{\ddag}$ & 0.420$^{\ddag}$ \\
          & w/o CVL & \textbf{0.050} & 0.899 & 0.768 & 0.739 & 0.511 & 0.544 \\
          & \textbf{InfoCTM} & 0.048 & \textbf{0.913} & 0.760 & \textbf{0.747} & \textbf{0.545} & \textbf{0.556} \\
    \bottomrule
    \end{tabular}%
    \caption{
        Experiment with low-coverage dictionaries and ablation study.
        Here w/o CVL means InfoCTM without the cross-lingual vocabulary linking method and only uses the translation pairs from a dictionary as linked words.
        The superscript $\ddag$ denotes that the improvements of InfoCTM are statistically significant at 0.05 level.
    }
    \label{tab_dict_size}%
  \end{table}%

    \subsection{Intra-lingual and Cross-lingual Classification}
        As mentioned previously,
        doc-topic distributions of a cross-lingual topic model should be cross-lingually consistent
        and provide transferable features for cross-lingual tasks.
        To evaluate this performance,
        we train SVM classifiers with doc-topic distributions as features and compare their accuracy with F1 scores
        following \citet{yuan2018multilingual}.
        Specifically, we evaluate the classification performance from two perspectives.
        \begin{inparaenum}[(i)]
            \item
                \textbf{Intra-lingual} classification (\abr{\textbf{-i}}): we train and test the classifier on the \emph{same} language.
            \item
                \textbf{Cross-lingual} classification (\abr{\textbf{-c}}): we train the classifier on one language and test it on another.
        \end{inparaenum}
        For example,
        Amazon Review dataset includes English (\abr{en}) and Chinese (\abr{zh}) documents;
        ``\abr{zh-i}'' denotes the classifier is trained and tested both on Chinese,
        while ``\textsc{zh-c}'' denotes the classifier is trained on English and tested on Chinese.

        As shown in \Cref{fig_classification},
        the intra-lingual classification accuracy (\abr{en-i,zh-i,ja-i}) of InfoCTM is much higher than MCTA and MTAnchor, and is close to NMTM.
        This is reasonable since InfoCTM and NMTM both infer doc-topic distributions in the framework of VAE.
        Nevertheless, InfoCTM achieves clearly higher cross-lingual classification accuracy (\abr{en-c,zh-c,ja-c}),
        and the improvements are statistically significant at 0.05 level.
        The reason lies in that InfoCTM uses our proposed topic alignment with mutual information method instead of the direct alignment of NMTM.
        This new method enhances topic alignment across languages 
        and
        thus produces more consistent and transferable doc-topic distributions than NMTM.
        In a word, these results show that InfoCTM has better transferability for cross-lingual classification tasks.

    \subsection{Low-coverage Dictionary and Ablation Study}
        To evaluate the performance with low-coverage dictionaries,
        we experiment with different dictionary sizes following \citet{hao2018learning}.
        Meanwhile, we conduct an ablation study on the proposed cross-lingual vocabulary linking (CVL) method.
        Let w/o CVL denote InfoCTM but without CVL and using the translation pairs from dictionaries as linked words only.
        \Cref{tab_dict_size} reports the topic quality and classification results
        under different dictionary sizes (25\%, 50\%, 75\%, and 100\%) on EC News.
        We only include NMTM in this study as NMTM outperforms all other baselines.

        We have the following observations from \Cref{tab_dict_size}:
        \begin{inparaenum}[(\bgroup\bfseries i\egroup)]
            \item
                InfoCTM can perform well with low-coverage dictionaries.
                Compared to NMTM, InfoCTM achieves better topic quality concerning CNPMI and $TU$.
                Similar to previous experiments,
                the intra-lingual accuracy (\abr{en-i}, \abr{zh-i}) of InfoCTM is close to NMTM,
                but its cross-lingual accuracy (\abr{en-c}, \abr{zh-c}) is obviously higher.
                We also see InfoCTM with 25\% of the dictionary achieves close performance to NMTM with 100\% of the dictionary.
            \item
                Our proposed CVL method effectively mitigates the low-coverage dictionary issue.
                InfoCTM and w/o CVL have similar CNPMI scores,
                but InfoCTM has increasingly higher $TU$ and cross-lingual accuracy along with smaller dictionary sizes.
                These show our CVL method can improve the performance
                when only low-coverage dictionaries are available.
            \end{inparaenum}

\begin{CJK*}{UTF8}{gbsn}
      \begin{table}[!t]
              \centering
              \small
              \setlength{\tabcolsep}{1pt}
              \renewcommand{\arraystretch}{1.1}

\begin{tabular}{rccccc}
    \toprule
    & \multicolumn{5}{c}{Top related words of topics} \\
    \midrule
    \multicolumn{6}{c}{NMTM} \\
            {\textbf{\abr{en} Topic\#1:}} & \underline{sport} & \underline{thrones} & soccer & \underline{episode} & bachelor \\
            {\textbf{\abr{zh} Topic\#1:}} & \underline{球队} & \underline{球员} & \underline{球迷} & 巴萨    & \underline{穆里尼奥} \\
            \emph{translations}:      &  \emph{team} & \emph{player} & \emph{fans} & \emph{abrcelona} & \emph{mourinho} \\
    \cmidrule{2-6}
          {\textbf{\abr{en} Topic\#2:}} & \underline{sport} & \underline{thrones} & \underline{episode} & hes   & wars \\
          {\textbf{\abr{zh} Topic\#2:}} & \underline{球队} & \underline{球迷} & \underline{球员} & \underline{穆里尼奥} & 皇马 \\
          \emph{translations}:      & \emph{team} & \emph{fans} & \emph{player} & \emph{mourinho} & \emph{real adrid}) \\
    \midrule
    \multicolumn{6}{c}{\textbf{InfoCTM}} \\
          {\textbf{\abr{en} Topic\#1:}} & club  & rent  & abrcelona & milan & chelsea \\
          {\textbf{\abr{zh} Topic\#1:}} & 转会    & 租借    & 米兰    & 俱乐部   & 切尔西 \\
          \emph{translations}:     & \emph{transfer} & \emph{rent} & \emph{milan} & \emph{club} & \emph{chelsea} \\
    \midrule
    \midrule
    \multicolumn{6}{c}{NMTM} \\
          {\textbf{\abr{en} Topic\#1:}} & learn & book  & sweat & teach & exercise \\
          {\textbf{\abr{jp} Topic\#1:}} & \begin{CJK*}{UTF8}{bsmi}愛\end{CJK*}用    & 年     & 使い    & 助かり   & シャンプ \\
          \emph{translations}:      & \emph{favorite} & \emph{year} & \emph{use} & \emph{help} & \emph{shampoo} \\
    \midrule
    \multicolumn{6}{c}{\textbf{InfoCTM}} \\
          {\textbf{\abr{en} Topic\#1:}} & yoga  & workout & exercise & drinking & drink \\
          {\textbf{\abr{jp} Topic\#1:}} & ヨガ    & \begin{CJK*}{UTF8}{bsmi}飲\end{CJK*}ん    & \begin{CJK*}{UTF8}{bsmi}運動\end{CJK*}    & \begin{CJK*}{UTF8}{bsmi}飲\end{CJK*}み物   & 肌 \\
          \emph{translations}:      & \emph{yoga} & \emph{drinking} & \emph{exercise} & \emph{drinking} & \emph{body} \\
    \bottomrule
    \end{tabular}%
                  \caption{
                      Top related words of discovered topics in each row.
                      Repetitive words are underlined.
                      \emph{Words} in italics are the translations of the above Chinese or Japanese words.
                  }
                  \label{tab_topic_examples}%
      \end{table}%
\end{CJK*}

    \subsection{Case Study of Discovered Topics} \label{sec_examples_topics}
        To qualitatively evaluate the topic quality,
        we conduct a case study of discovered topics selected by querying keywords ``soccer'' and ``exercise''.
        They are shown in \Cref{tab_topic_examples} (translations are in the brackets for easier understanding, and they are \emph{not} the words of discovered topics).
        Recall that well-aligned topics should be semantically consistent across languages.
        For the topic ``soccer'' from EC News,
        NMTM produces repetitive topics with repeated words like ``sports'' and ``episode''.
        In contrast, InfoCTM only generates one relevant topic about soccer and the words are clearly coherent with the words ``club'', ``milan'', and ``chelsea''.
        For the topic ``exercise'' from Rakuten Amazon,
        InfoCTM aligns the topics well with relevant words in English and Japanese, \emph{e.g.}, ``yoga'', ``exercise'' and ``drinking''.
        But NMTM wrongly aligns the topics with irrelevant and incoherent words.

    \subsection{Visualization of Latent Space}
        We use t-SNE \cite{Maaten2008} to visualize the learned topic representations of the top related words of English and Chinese topics discovered by our InfoCTM from EC News.
        The Appendix
        shows
        the topics are well-aligned across languages,
        and the topic representations of words are well-grouped and separately scattered
        in the latent space.
        For example, English Topic\#2 and Chinese Topic\#2 are both about music,
        including the words ``song'', ``album'', and ``sing''.
        These words are close to each other while away from words of other topics.
        We notice Topic\#2 about music and Topic\#3 about the movie are closer to each other on the canvas as they are relatively more related.
        This qualitatively verifies our InfoCTM indeed properly aligns the topic representations and prevents degenerate topic representations.

\section{Conclusion}
    In this paper,
    we propose InfoCTM to discover aligned latent topics of cross-lingual corpora.
    InfoCTM uses the novel topic alignment with mutual information method that avoids the repetitive topic issue and uses a new cross-lingual vocabulary linking method that alleviates the low-coverage issue.
    Experiments show that InfoCTM can consistently outperform baselines,
    producing higher-quality topics and showing better transferability for cross-lingual downstream tasks.
    Especially, InfoCTM can perform well under low-coverage dictionaries, making it applicable for more scenarios like low-resource languages.

\section*{Acknowledgements}
    We want to thank all anonymous reviewers for their helpful comments.
    This research/project is supported by the National Research Foundation, Singapore under its AI Singapore Programme, AISG Award No: AISG2-TC-2022-005.
\bibliography{lib}

\begin{thebibliography}{50}
\providecommand{\natexlab}[1]{#1}

\bibitem[{Arora et~al.(2019)Arora, Khandeparkar, Khodak, Plevrakis, and
  Saunshi}]{arora2019theoretical}
Arora, S.; Khandeparkar, H.; Khodak, M.; Plevrakis, O.; and Saunshi, N. 2019.
\newblock A theoretical analysis of contrastive unsupervised representation
  learning.
\newblock In \emph{36th International Conference on Machine Learning, ICML
  2019}, 9904--9923. International Machine Learning Society (IMLS).

\bibitem[{Bachman, Hjelm, and Buchwalter(2019)}]{bachman2019learning}
Bachman, P.; Hjelm, R.~D.; and Buchwalter, W. 2019.
\newblock Learning representations by maximizing mutual information across
  views.
\newblock \emph{Advances in neural information processing systems}, 32.

\bibitem[{Bianchi et~al.(2020)Bianchi, Terragni, Hovy, Nozza, and
  Fersini}]{bianchi2020cross}
Bianchi, F.; Terragni, S.; Hovy, D.; Nozza, D.; and Fersini, E. 2020.
\newblock Cross-lingual contextualized topic models with zero-shot learning.
\newblock \emph{arXiv preprint arXiv:2004.07737}.

\bibitem[{Blei and Lafferty(2006)}]{blei2006dynamic}
Blei, D.~M.; and Lafferty, J.~D. 2006.
\newblock Dynamic topic models.
\newblock In \emph{Proceedings of the 23rd international conference on Machine
  learning}, 113--120.

\bibitem[{Blei, Ng, and Jordan(2003)}]{blei2003latent}
Blei, D.~M.; Ng, A.~Y.; and Jordan, M.~I. 2003.
\newblock {Latent dirichlet allocation}.
\newblock \emph{Journal of Machine Learning Research}, 3(Jan): 993--1022.

\bibitem[{Boyd-Graber and Blei(2012)}]{boyd2012multilingual}
Boyd-Graber, J.; and Blei, D. 2012.
\newblock Multilingual topic models for unaligned text.
\newblock \emph{arXiv preprint arXiv:1205.2657}.

\bibitem[{Chang and Hwang(2021)}]{chang2021word}
Chang, C.-H.; and Hwang, S.-Y. 2021.
\newblock A word embedding-based approach to cross-lingual topic modeling.
\newblock \emph{Knowledge and Information Systems}, 63(6): 1529--1555.

\bibitem[{Chang et~al.(2009)Chang, Gerrish, Wang, Boyd-Graber, and
  Blei}]{chang2009reading}
Chang, J.; Gerrish, S.; Wang, C.; Boyd-Graber, J.~L.; and Blei, D.~M. 2009.
\newblock {Reading tea leaves: How humans interpret topic models}.
\newblock In \emph{Advances in neural information processing systems},
  288--296.

\bibitem[{Chen et~al.(2020)Chen, Kornblith, Norouzi, and
  Hinton}]{chen2020simple}
Chen, T.; Kornblith, S.; Norouzi, M.; and Hinton, G. 2020.
\newblock A simple framework for contrastive learning of visual
  representations.
\newblock In \emph{International conference on machine learning}, 1597--1607.
  PMLR.

\bibitem[{Chi et~al.(2020)Chi, Dong, Wei, Yang, Singhal, Wang, Song, Mao,
  Huang, and Zhou}]{chi2020infoxlm}
Chi, Z.; Dong, L.; Wei, F.; Yang, N.; Singhal, S.; Wang, W.; Song, X.; Mao,
  X.-L.; Huang, H.; and Zhou, M. 2020.
\newblock InfoXLM: An information-theoretic framework for cross-lingual
  language model pre-training.
\newblock \emph{arXiv preprint arXiv:2007.07834}.

\bibitem[{Conneau et~al.(2017)Conneau, Lample, Ranzato, Denoyer, and
  J{\'e}gou}]{conneau2017word}
Conneau, A.; Lample, G.; Ranzato, M.; Denoyer, L.; and J{\'e}gou, H. 2017.
\newblock Word Translation Without Parallel Data.
\newblock \emph{arXiv preprint arXiv:1710.04087}.

\bibitem[{Devlin et~al.(2018)Devlin, Chang, Lee, and
  Toutanova}]{devlin2018bert}
Devlin, J.; Chang, M.-W.; Lee, K.; and Toutanova, K. 2018.
\newblock Bert: Pre-training of deep bidirectional transformers for language
  understanding.
\newblock \emph{arXiv preprint arXiv:1810.04805}.

\bibitem[{Dong et~al.(2021)Dong, Luu, Lin, Yan, and Zhang}]{dong2021should}
Dong, X.; Luu, A.~T.; Lin, M.; Yan, S.; and Zhang, H. 2021.
\newblock How Should Pre-Trained Language Models Be Fine-Tuned Towards
  Adversarial Robustness?
\newblock \emph{Advances in Neural Information Processing Systems}, 34:
  4356--4369.

\bibitem[{Guti{\'e}rrez et~al.(2016)Guti{\'e}rrez, Shutova, Lichtenstein,
  de~Melo, and Gilardi}]{gutierrez2016detecting}
Guti{\'e}rrez, E.~D.; Shutova, E.; Lichtenstein, P.; de~Melo, G.; and Gilardi,
  L. 2016.
\newblock Detecting cross-cultural differences using a multilingual topic
  model.
\newblock \emph{Transactions of the Association for Computational Linguistics},
  4: 47--60.

\bibitem[{Hao, Boyd-Graber, and Paul(2018)}]{hao2018lessons}
Hao, S.; Boyd-Graber, J.~L.; and Paul, M.~J. 2018.
\newblock Lessons from the bible on modern topics: adapting topic model
  evaluation to multilingual and low-resource settings.
\newblock In \emph{Proceedings of the 2018 conference of the North American
  chapter of the association for computational linguistics: human language
  technologies, NAACL-HLT}, 1--6.

\bibitem[{Hao and Paul(2018)}]{hao2018learning}
Hao, S.; and Paul, M. 2018.
\newblock Learning multilingual topics from incomparable corpora.
\newblock In \emph{Proceedings of the 27th international conference on
  computational linguistics}, 2595--2609.

\bibitem[{Hao and Paul(2020)}]{hao2020empirical}
Hao, S.; and Paul, M.~J. 2020.
\newblock An empirical study on crosslingual transfer in probabilistic topic
  models.
\newblock \emph{Computational Linguistics}, 46(1): 95--134.

\bibitem[{Hjelm et~al.(2019)Hjelm, Fedorov, Lavoie-Marchildon, Grewal, Bachman,
  Trischler, and Bengio}]{hjelm2018learning}
Hjelm, R.~D.; Fedorov, A.; Lavoie-Marchildon, S.; Grewal, K.; Bachman, P.;
  Trischler, A.; and Bengio, Y. 2019.
\newblock Learning deep representations by mutual information estimation and
  maximization.
\newblock In \emph{International Conference on Learning Representations}.

\bibitem[{Jagarlamudi and Daum{\'e}(2010)}]{jagarlamudi2010extracting}
Jagarlamudi, J.; and Daum{\'e}, H. 2010.
\newblock Extracting multilingual topics from unaligned comparable corpora.
\newblock In \emph{European Conference on Information Retrieval}, 444--456.
  Springer.

\bibitem[{Kingma and Welling(2014)}]{Kingma2014a}
Kingma, D.~P.; and Welling, M. 2014.
\newblock {Auto-encoding variational bayes}.
\newblock In \emph{The International Conference on Learning Representations
  (ICLR)}.

\bibitem[{Koehn(2005)}]{koehn2005europarl}
Koehn, P. 2005.
\newblock Europarl: A parallel corpus for statistical machine translation.
\newblock In \emph{Proceedings of machine translation summit x: papers},
  79--86.

\bibitem[{Kong et~al.(2020)Kong, de~Masson~d'Autume, Yu, Ling, Dai, and
  Yogatama}]{kong2020}
Kong, L.; de~Masson~d'Autume, C.; Yu, L.; Ling, W.; Dai, Z.; and Yogatama, D.
  2020.
\newblock A Mutual Information Maximization Perspective of Language
  Representation Learning.
\newblock In \emph{International Conference on Learning Representations}.

\bibitem[{Lind et~al.(2019)Lind, Eberl, Galyga, Heidenreich, Boomgaarden,
  Jim{\'e}nez, and Berganza}]{lind2019bridge}
Lind, F.; Eberl, J.-M.; Galyga, S.; Heidenreich, T.; Boomgaarden, H.~G.;
  Jim{\'e}nez, B.~H.; and Berganza, R. 2019.
\newblock A bridge over the language gap: Topic modelling for text analyses
  across languages for country comparative research.
\newblock \emph{University of Vienna: Working Paper of the REMINDER-Project}.

\bibitem[{Logeswaran and Lee(2018)}]{logeswaran2018efficient}
Logeswaran, L.; and Lee, H. 2018.
\newblock An efficient framework for learning sentence representations.
\newblock In \emph{International Conference on Learning Representations}.

\bibitem[{Miao, Yu, and Blunsom(2016)}]{Miao2016}
Miao, Y.; Yu, L.; and Blunsom, P. 2016.
\newblock {Neural variational inference for text processing}.
\newblock In \emph{International conference on machine learning}, 1727--1736.

\bibitem[{Mikolov et~al.(2013)Mikolov, Chen, Corrado, and Dean}]{mikolov2013}
Mikolov, T.; Chen, K.; Corrado, G.~S.; and Dean, J. 2013.
\newblock Efficient Estimation of Word Representations in Vector Space.
\newblock In \emph{ICLR}.

\bibitem[{Mimno et~al.(2009)Mimno, Wallach, Naradowsky, Smith, and
  McCallum}]{mimno2009polylingual}
Mimno, D.; Wallach, H.; Naradowsky, J.; Smith, D.~A.; and McCallum, A. 2009.
\newblock Polylingual topic models.
\newblock In \emph{Proceedings of the 2009 conference on empirical methods in
  natural language processing}, 880--889.

\bibitem[{Mueller and Dredze(2021)}]{mueller2021fine}
Mueller, A.; and Dredze, M. 2021.
\newblock Fine-tuning Encoders for Improved Monolingual and Zero-shot
  Polylingual Neural Topic Modeling.
\newblock In \emph{Proceedings of the 2021 Conference of the North American
  Chapter of the Association for Computational Linguistics: Human Language
  Technologies}, 3054--3068.

\bibitem[{Nan et~al.(2019)Nan, Ding, Nallapati, and Xiang}]{Nan2019}
Nan, F.; Ding, R.; Nallapati, R.; and Xiang, B. 2019.
\newblock Topic Modeling with {W}asserstein Autoencoders.
\newblock In \emph{Proceedings of the 57th Annual Meeting of the Association
  for Computational Linguistics}, 6345--6381. Florence, Italy: Association for
  Computational Linguistics.

\bibitem[{Newman et~al.(2010)Newman, Lau, Grieser, and Baldwin}]{Newman2010}
Newman, D.; Lau, J.~H.; Grieser, K.; and Baldwin, T. 2010.
\newblock {Automatic evaluation of topic coherence}.
\newblock In \emph{Human Language Technologies: The 2010 Annual Conference of
  the North American Chapter of the Association for Computational Linguistics},
  100--108. Association for Computational Linguistics.
\newblock ISBN 1932432655.

\bibitem[{Nguyen et~al.(2022)Nguyen, Wu, Luu, Nguyen, Hai, and
  Bing}]{nguyen2022adaptive}
Nguyen, T.; Wu, X.; Luu, A.-T.; Nguyen, C.-D.; Hai, Z.; and Bing, L. 2022.
\newblock Adaptive Contrastive Learning on Multimodal Transformer for Review
  Helpfulness Predictions.
\newblock \emph{arXiv preprint arXiv:2211.03524}.

\bibitem[{Ni et~al.(2009)Ni, Sun, Hu, and Chen}]{ni2009mining}
Ni, X.; Sun, J.-T.; Hu, J.; and Chen, Z. 2009.
\newblock Mining multilingual topics from Wikipedia.
\newblock In \emph{Proceedings of the 18th international conference on World
  wide web}, 1155--1156.

\bibitem[{Shi et~al.(2016)Shi, Lam, Bing, and Xu}]{shi2016detecting}
Shi, B.; Lam, W.; Bing, L.; and Xu, Y. 2016.
\newblock Detecting common discussion topics across culture from news reader
  comments.
\newblock In \emph{Proceedings of the 54th Annual Meeting of the Association
  for Computational Linguistics (Volume 1: Long Papers)}, 676--685.

\bibitem[{Srivastava and Sutton(2017)}]{Srivastava2017}
Srivastava, A.; and Sutton, C. 2017.
\newblock {Autoencoding variational inference for topic models}.
\newblock In \emph{ICLR}.

\bibitem[{Tian, Krishnan, and Isola(2020)}]{tian2020contrastive}
Tian, Y.; Krishnan, D.; and Isola, P. 2020.
\newblock Contrastive multiview coding.
\newblock In \emph{European conference on computer vision}, 776--794. Springer.

\bibitem[{Van~den Oord, Li, and Vinyals(2018)}]{van2018representation}
Van~den Oord, A.; Li, Y.; and Vinyals, O. 2018.
\newblock Representation learning with contrastive predictive coding.
\newblock \emph{arXiv e-prints}, arXiv--1807.

\bibitem[{van~der Maaten and Hinton(2008)}]{Maaten2008}
van~der Maaten, L.; and Hinton, G. 2008.
\newblock {Visualizing data using t-SNE}.
\newblock \emph{Journal of machine learning research}, 9(Nov): 2579--2605.

\bibitem[{Vuli{\'c}, De~Smet, and Moens(2013)}]{vulic2013cross}
Vuli{\'c}, I.; De~Smet, W.; and Moens, M.-F. 2013.
\newblock Cross-language information retrieval models based on latent topic
  models trained with document-aligned comparable corpora.
\newblock \emph{Information Retrieval}, 16(3): 331--368.

\bibitem[{Wang and Isola(2020)}]{wang2020understanding}
Wang, T.; and Isola, P. 2020.
\newblock Understanding contrastive representation learning through alignment
  and uniformity on the hypersphere.
\newblock In \emph{International Conference on Machine Learning}, 9929--9939.
  PMLR.

\bibitem[{Wu et~al.(2022)Wu, Dong, Nguyen, and Luu}]{wu2022}
Wu, X.; Dong, X.; Nguyen, T.~T.; and Luu, A.~T. 2022.
\newblock Neural Topic Modeling with Embedding Clustering Regularization.
\newblock Forthcoming.

\bibitem[{Wu and Li(2019)}]{Wu2019}
Wu, X.; and Li, C. 2019.
\newblock {Short Text Topic Modeling with Flexible Word Patterns}.
\newblock In \emph{International Joint Conference on Neural Networks}.

\bibitem[{Wu, Li, and Miao(2021)}]{wu2021discovering}
Wu, X.; Li, C.; and Miao, Y. 2021.
\newblock Discovering Topics in Long-tailed Corpora with Causal Intervention.
\newblock In \emph{Findings of the Association for Computational Linguistics:
  ACL-IJCNLP 2021}, 175--185. Online: Association for Computational
  Linguistics.

\bibitem[{Wu et~al.(2020{\natexlab{a}})Wu, Li, Zhu, and Miao}]{Wu2020}
Wu, X.; Li, C.; Zhu, Y.; and Miao, Y. 2020{\natexlab{a}}.
\newblock {Learning Multilingual Topics with Neural Variational Inference}.
\newblock In \emph{International Conference on Natural Language Processing and
  Chinese Computing}.

\bibitem[{Wu et~al.(2020{\natexlab{b}})Wu, Li, Zhu, and Miao}]{Wu2020short}
Wu, X.; Li, C.; Zhu, Y.; and Miao, Y. 2020{\natexlab{b}}.
\newblock Short Text Topic Modeling with Topic Distribution Quantization and
  Negative Sampling Decoder.
\newblock In \emph{Proceedings of the 2020 Conference on Empirical Methods in
  Natural Language Processing (EMNLP)}, 1772--1782. Online.

\bibitem[{Wu, Luu, and Dong(2022)}]{Wu2022mitigating}
Wu, X.; Luu, A.~T.; and Dong, X. 2022.
\newblock Mitigating Data Sparsity for Short Text Topic Modeling by
  Topic-Semantic Contrastive Learning.
\newblock arXiv:2211.12878.

\bibitem[{Wu et~al.(2018)Wu, Xiong, Yu, and Lin}]{wu2018unsupervised}
Wu, Z.; Xiong, Y.; Yu, S.~X.; and Lin, D. 2018.
\newblock Unsupervised feature learning via non-parametric instance
  discrimination.
\newblock In \emph{Proceedings of the IEEE conference on computer vision and
  pattern recognition}, 3733--3742.

\bibitem[{Xu et~al.(2022)Xu, Lu, Li, Wu, Qi, Ye, Wang, and Zhou}]{xu2022neural}
Xu, K.; Lu, X.; Li, Y.-f.; Wu, T.; Qi, G.; Ye, N.; Wang, D.; and Zhou, Z. 2022.
\newblock Neural Topic Modeling with Deep Mutual Information Estimation.
\newblock \emph{arXiv preprint arXiv:2203.06298}.

\bibitem[{Yang, Boyd-Graber, and Resnik(2019)}]{yang2019multilingual}
Yang, W.; Boyd-Graber, J.; and Resnik, P. 2019.
\newblock A multilingual topic model for learning weighted topic links across
  corpora with low comparability.
\newblock In \emph{Proceedings of the 2019 Conference on Empirical Methods in
  Natural Language Processing and the 9th International Joint Conference on
  Natural Language Processing (EMNLP-IJCNLP)}, 1243--1248.

\bibitem[{Yuan, Van~Durme, and Ying(2018)}]{yuan2018multilingual}
Yuan, M.; Van~Durme, B.; and Ying, J.~L. 2018.
\newblock Multilingual anchoring: Interactive topic modeling and alignment
  across languages.
\newblock \emph{Advances in neural information processing systems}, 31.

\bibitem[{Zhang and LeCun(2017)}]{zhang2017encoding}
Zhang, X.; and LeCun, Y. 2017.
\newblock Which encoding is the best for text classification in Chinese,
  English, Japanese and Korean?
\newblock \emph{arXiv preprint arXiv:1708.02657}.

\end{thebibliography}

\clearpage
\appendix
\section{Appendix}
\subsection{Visualization of Latent Space} \label{sec_app_visualization}
\begin{figure}[!ht]
    \centering
    \includegraphics[width=\linewidth]{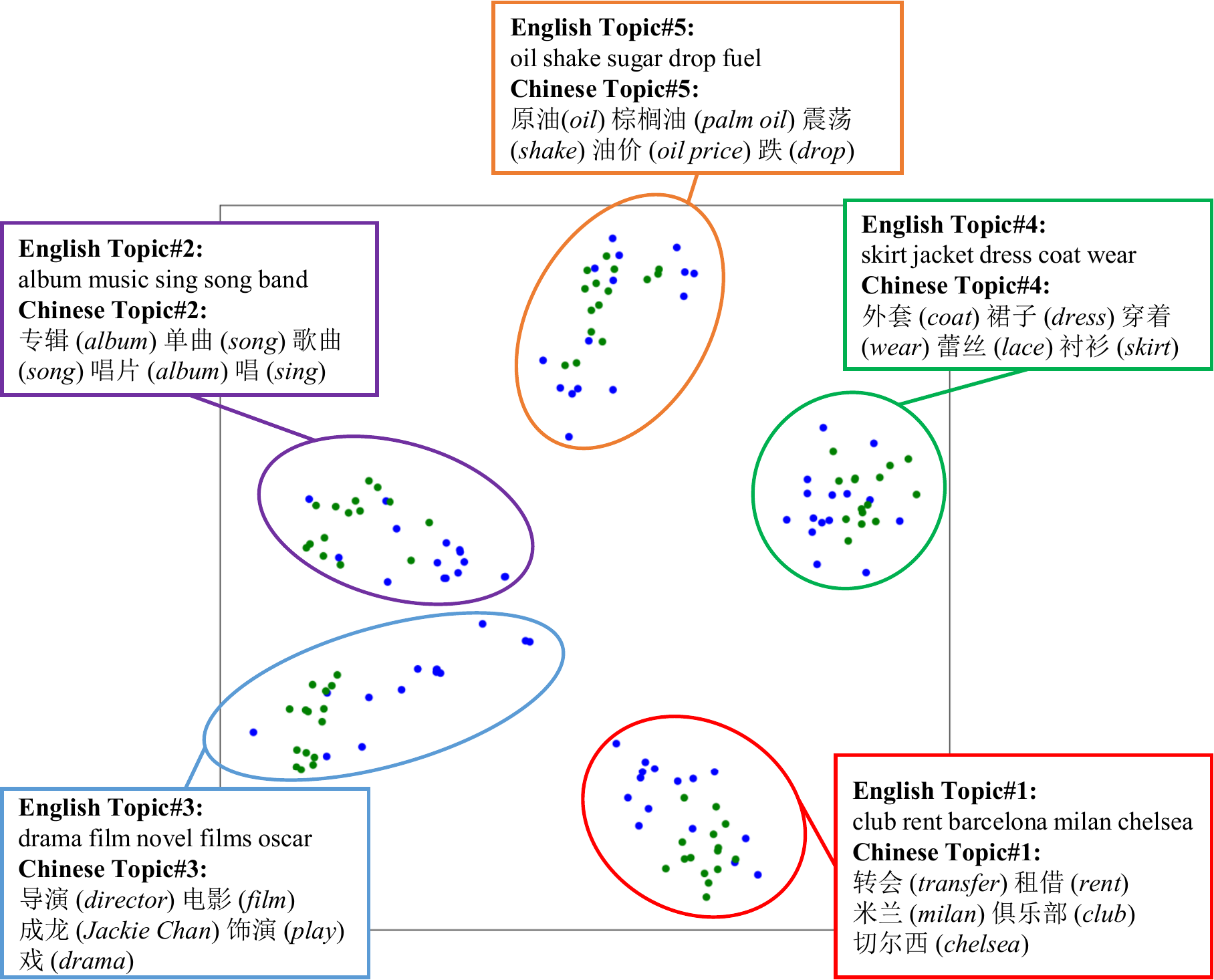}
    \caption{
        t-SNE visualization of learned topic representations of words.
        Blue points denote English words and green points Chinese words.
        \emph{Words} in the brackets are translations.
    }
    \label{fig_topic_representation}
\end{figure}

\end{document}